\documentclass[letterpaper]{article} 
\usepackage[submission]{paper}  
\usepackage{times}  
\usepackage{helvet}  
\usepackage{courier}  
\usepackage[hyphens]{url}  
\usepackage{graphicx} 
\usepackage{natbib}  
\usepackage{caption} 
\usepackage{algorithm}
\usepackage{algorithmic}
\usepackage{float}

\usepackage{amsfonts}
\usepackage{amsmath}
\usepackage{color}
\usepackage{multirow}
\usepackage{booktabs}
\usepackage{tabularray}
\usepackage{newfloat}
\usepackage{booktabs}
\usepackage{listings}
\DeclareCaptionStyle{ruled}{labelfont=normalfont,labelsep=colon,strut=off} 
\floatstyle{ruled}
\newfloat{listing}{tb}{lst}{}
\floatname{listing}{Listing}

\newcommand{\minisection}[1]{\vspace{0.02in}
\noindent {\bf #1}\ }

\title{Class Incremental Learning with Pre-trained Vision-Language Models}

\author{
Xialei Liu $^{1*}$ \quad
Xusheng Cao $^1$\thanks{The first two authors contribute equally.} \quad
Haori Lu $^1$ \quad
Jia-wen Xiao $^1$ \quad \\
Andrew D. Bagdanov $^{2}$ \quad
Ming-Ming Cheng $^1$ \\
$^1$ VCIP, CS, Nankai University \quad
$^2$  MICC, University of Florence \\
~\\
\normalsize \{xialei,cmm\}@nankai.edu.cn, \quad andrew.bagdanov@unifi.it\\
\{caoxusheng, luhaori, xiaojw\}@mail.nankai.edu.cn 
}

\affiliations{
    {xialei, cmm}@nankai.edu.cn, {caoxusheng, luhaori, xiaojw}@mail.nankai.edu.cn, andrew.bagdanov@unifi.it
}

\usepackage{bibentry}

\begin{document}

\maketitle

\begin{abstract}
With the advent of large-scale pre-trained models, interest in adapting and exploiting them for continual learning scenarios has grown. 
   In this paper, we propose an approach to exploiting pre-trained vision-language models (e.g. CLIP) that enables further adaptation instead of only using zero-shot learning of new tasks. We augment a pre-trained CLIP model with additional layers after the Image Encoder or before the Text Encoder. We investigate three different strategies: a Linear Adapter, a Self-attention Adapter, each operating on the image embedding, and Prompt Tuning which instead modifies prompts input to the CLIP text encoder. We also propose a method for parameter retention in the adapter layers that uses a measure of parameter importance to better maintain stability and plasticity during incremental learning. Our experiments demonstrate that the simplest solution -- a single Linear Adapter layer with parameter retention -- produces the best results. Experiments on several conventional benchmarks consistently show a significant margin of improvement over the current state-of-the-art.
\end{abstract}

\section{Introduction}

Deep neural networks have revolutionized many real-world computer vision applications. However, most neural networks are restricted to static world scenarios in which they are trained \textit{once} and lack the flexibility to \textit{adapt and update} themselves in dynamically changing environments. The objective of continual learning 
is to enable training of neural networks in the presence of newly-arriving data~\cite{belouadah2020comprehensive,delange2021continual,masana2020class,parisi2019continual}. The key problem to overcome in continual learning is catastrophic forgetting in which the network, when integrating knowledge needed for solving new tasks, forgets knowledge key to solving previous ones~\cite{mccloskey1989catastrophic}.
Class incremental learning (CIL), in which tasks consist of disjoint classes, is one of the principal continual learning scenarios considered in the literature~\cite{van2019three}.  
 \begin{figure}[t]
\includegraphics[width=0.47\textwidth]{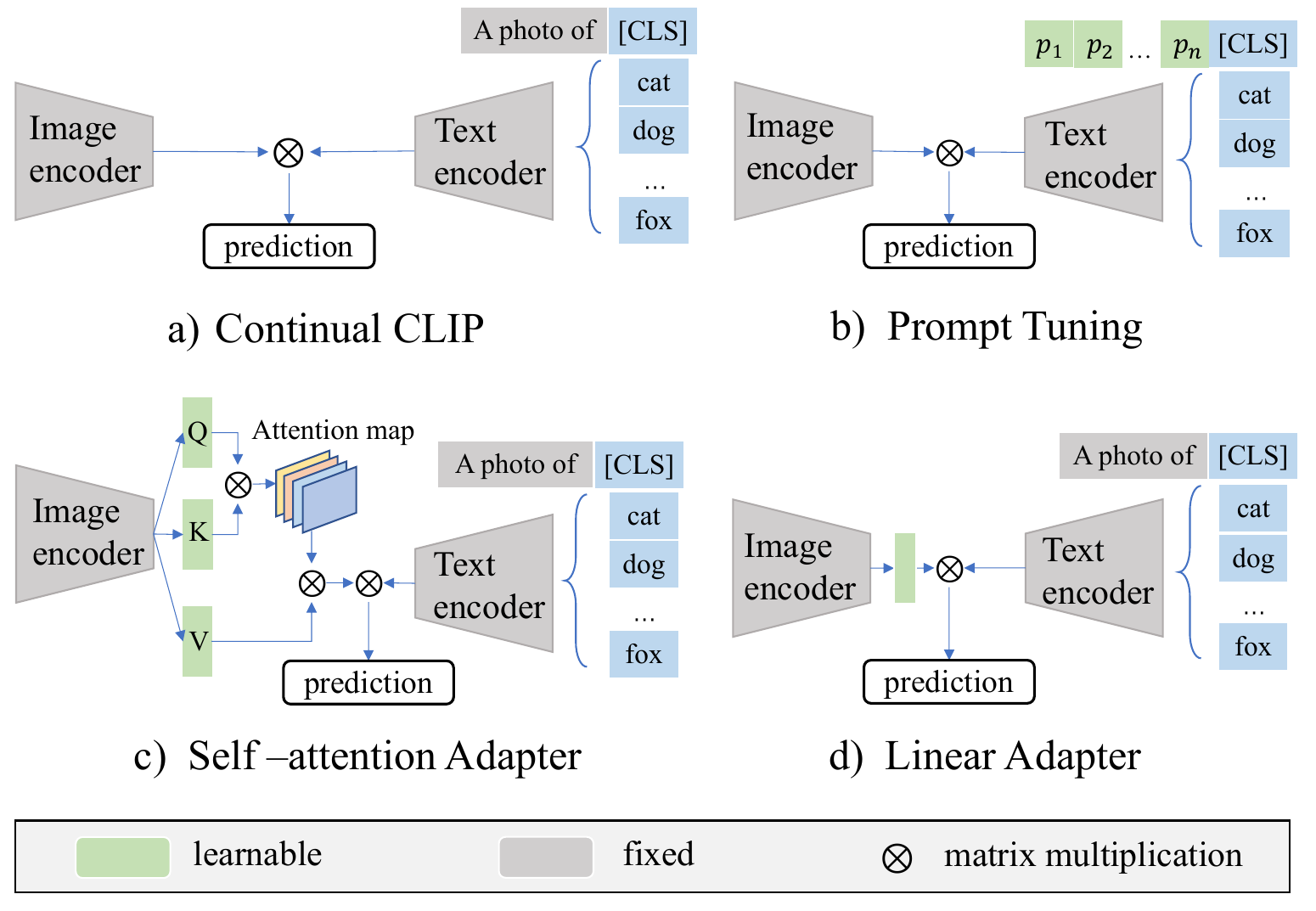}
\vspace{-8pt}
\caption{Different adaptation options for class incremental learning with pre-trained vision-language models. a) Zero-shot learning using CLIP for class incremental learning proposed in Continual-CLIP~\cite{continual-clip}. b) Prompt Tuning by learning additional text prompts. c) Self-attention Adapter for incremental adaptation of encoded image features. d) Linear Adapter after the image encoder for incremental adaptation of encoded image features. }
\label{fig:abs}
\end{figure}

In CIL, classes are presented to the learner in a sequence of disjoint \textit{tasks} containing one or more classes. CIL approaches can be roughly categorized into three broad classes: regularization-based, architecture-based, and replay-based strategies~\cite{delange2021continual}.
Replay-based CIL methods, which currently dominate the state-of-the-art, store a small set of samples (called exemplars) from previous tasks. Exemplars help consolidate previous knowledge, but they bring with them the new challenge of (often extreme) imbalance between current task data and exemplars of past tasks. This imbalance leads to predictions biased toward new classes, which in turn leads to worse performance on old ones. This problem is specifically addressed in IL2M~\cite{belouadah2019il2m}, UCIR~\cite{hou2019learning}, BiC~\cite{wu2019large}, EEIL~\cite{castro2018end}and SS-IL~\cite{ahn2021ss}.

A recent trend in CIL is the exploitation of pre-trained models to leverage the knowledge contained therein~\cite{ermis2022memory,l2p,wu2022class,xue2022meta}. The use of foundation models has yielded very strong performance gains in several domains. Most previous works are based on pre-trained vision models, for example, models pre-trained on ImageNet using supervised learning~\cite{wang2022dualprompt,l2p,xue2022meta}, or unsupervised learning~\cite{cossu2022continual,wang2022dualprompt,l2p}. Continual-CLIP~\cite{continual-clip} is the first work evaluating the potential of exploiting a pre-trained vision-language (CLIP~\cite{clip}) for continual learning. Continual-CLIP provides a very strong and simple baseline that exploits the zero-shot learning capabilities of CLIP, although the pre-trained model is frozen and is thus unable to further improve with new data.
 
In this work, we propose a simple and efficient paradigm for continual learning based on the adaptation of a pre-trained CLIP model. Specifically, during training, we extract the latent feature of each input image using the pre-trained CLIP Image Encoder. To further improve the model when new data arrives, we analyze three different ways of adding additional \emph{adaptation} parameters: Prompt Tuning, Self-attention Adapters, and Linear Adapters  (illustrated in Figure~\ref{fig:abs}). Adapted features can be further combined with the text features obtained with the CLIP Text Encoder. The inputs to the Text Encoder are hand-crafted prompts representing all categories the model has seen so far (e.g. ``A photo of a dog''). We follow the default approach of CLIP and multiply the feature matrices of the two modalities to obtain the final logits for classification. We then calculate the loss and perform backpropagation to update the parameters of the adapter. Note that the \emph{adapter} is the only component that needs to be updated, as we freeze all other parameters in CLIP. The adapter is temporarily retained and used for the initialization of the next task. 

Additionally, to better preserve the knowledge from the previous task, we propose a strategy for parameter retention in order to preserve important weights from the previous adapter. Only the most important parameters are updated, where importance measured by the distance between the previous and current parameters. We compare our method with conventional knowledge distillation approaches, and our experiments demonstrate the superiority of our method for preserving the knowledge of previous tasks.

To summarize, the main contributions of this work are:
\begin{itemize}
    \item we exploit a pre-trained CLIP model for CIL and integrate an additional \textit{adapter} to enable incorporation of new knowledge and adaptation to new tasks;
    \item we propose a parameter retention strategy to accumulate previous knowledge in a way that accounts for the importance of each parameter;
    \item we achieve state-of-the-art results on several datasets in different scenarios.
\end{itemize}

\section{Related Work}
We review work from the literature related to our approach.

\subsection{Class incremental learning}
Class incremental learning is one of the three most common settings in continual learning~\cite{van2019three}.

Approaches can be divided into three main categories. regularization-based,  parameter-isolation-based, and replay-based methods~\cite{delange2021continual}.  

\minisection{Regularization-based methods.}
The main idea behind regularization-based methods is to constrain the optimization of parameters by either identifying the importance of each or regularizing network outputs. Elastic Weight Consolidation (EWC)~\cite{kirkpatrick2017overcoming} uses a diagonal approximation to the Fisher Information Matrix (FIM) as a measure of parameter importance~\cite{kirkpatrick2017overcoming}. Rotated-EWC found the assumption of diagonal FIM to be too strict in practice and reparameterizes the network to approximately diagonalize the FIM by rotating the parameter space~\cite{liu2018rotate}. Synaptic Intelligence (SI)~\cite{zenke2017continual} and Memory Aware Synapses (MAS)~\cite{aljundi2018memory} are in this same research direction. Semantic Drift Compensation (SDC) is based on metric learning combined with regularization and semantic drift compensation~\cite{yu2020semantic}. There are also multiple works focusing on different variants of knowledge distillation~\cite{hu2021distilling, simon2021learning,tao2020topology} to alleviate forgetting.

\minisection{Architecture-based methods.}
Architecture-based methods manage model capacity by adding neurons and branches or by adding masks on top of the base model. Progressive Neural Networks (PNN)~\cite{rusu2016progressive}  add a duplicate layer with lateral connections, and DER~\cite{yan2021dynamically} dynamically adds network branches.  Progress \& Compress~\cite{schwarz2018progress} and AANETs~\cite{liu2021adaptive} use two different types of network blocks, one for stability and one for plasticity. Piggyback~\cite{mallya2018piggyback}, Packnet~\cite{mallya2018packnet}, HAT~\cite{serra2018overcoming}, and Ternary Feature Masks~\cite{masana2020ternary} are mask-based methods to consolidate previous knowledge. Random Path Selection~\cite{rajasegaran2019random} builds and selects random paths for different tasks without backpropagating to previously learned modules.

\minisection{Replay-based methods.}
Replay-based methods assume that a small buffer is available for storing knowledge from past tasks. They can store real samples (known as exemplars), synthetic samples produced by generative models, or intermediate feature representations (like DGR~\cite{shin2017continual} and REMIND~\cite{hayes2020remind}). Most replay-based methods store real samples from previous tasks. However, due to the limited memory available for previous samples, there is an imbalance problem when learning with current data and exemplars. This was partially addressed by EEIL~\cite{castro2018end}, BiC~\cite{wu2019large}, IL2M~\cite{belouadah2019il2m}, Rainbow Memory~\cite{bang2021rainbow} and SS-IL~\cite{ahn2021ss}. PODNet applies an efficient spatial distillation loss throughout the  model to distill  knowledge from previous tasks~\cite{douillard2020podnet}.
GDumb~\cite{prabhu2020gdumb} proposes a simple baseline by greedily storing samples and learning from scratch using samples only available in the memory. Verwimp et al.~\cite{verwimp2021rehearsal} revealed the limits and merits of rehearsal methods in a systematical manner.

\minisection{Pre-trained models for CIL.}
Pre-trained models can be utilized and obtain high performance in continual learning. Tz-Ying Wu et al.~\cite{wu2022class} propose a two-stage training strategy, cloning part of the pre-trained model and fine-tuning it on the new data in stage one, and combining the base and novel classifiers by a new fusion algorithm in stage two. L2P~\cite{l2p} introduces the idea of prompting in incremental learning and improves the performance by learning a prompt pool memory to instruct the pre-trained model. DualPrompt\cite{wang2022dualprompt} categorizes prompts as G-prompts and E-prompts which learn task-invariant and task-specific knowledge respectively. Although these two works achieved impressive results in a non-exemplar setting, it is worth noting that they utilized an ImageNet-21K pre-trained backbone. It is important to consider that the data distribution of ImageNet, CIFAR-100, and ImageNet-R is largely overlapping. Furthermore, \cite{janson2022simple}'s findings support the fact that using an ImageNet-21K pre-trained backbone with the NMC (non-parametric classification) approach yields results superior to L2P\cite{l2p} without the need for additional training. 

MEAT~\cite{xue2022meta} learns an attention mask for each task and dynamically assigns attention masks to generate task-specific self-attention patterns on a ViT~\cite{dosovitskiy2020image} backbone. 
ProgPrompt~\cite{razdaibiedina2023progressive} learns a new soft prompt for each task and sequentially concatenates it with previously learned prompts. ADA~\cite{ermis2022memory} proposes an adaptive distillation algorithm of adapters. It effectively consolidates new adapters with old adapters, so that pre-trained transformers can be trained incrementally on a sequence of tasks. In this work, we focus on class incremental learning with pre-trained vision-language models.

\subsection{Vision-Language Pre-training}
In recent years researchers have made great progress in both computer vision and natural language processing, and pre-trained vision-language models are attracting more and more attention. The pioneering work in CLIP~\cite{clip} proposes to directly learn from a large number of text-image pairs with a simple contrastive objective.

CoOp~\cite{CoOp} finds prompt engineering is a major challenge when deploying such models in practice, therefore it models text prompts with a learnable vector while the pre-trained parameters are kept fixed. CoCoOp~\cite{zhou2022conditional} further learns a lightweight neural network to generate an input-conditional token for each image to replace the static prompts of CoOp. The authors of ALIGN~\cite{jia2021scaling} found that previous models have strict requirements for datasets. They leverage a noisy dataset of over one billion image-text pairs and successfully use the scale of corpus to compensate for the noise.

The methods above are mainly suited for vision-based downstream tasks. To adapt to Vision-Language tasks, researchers propose to learn joint image-text embeddings, such as OSCAR~\cite{li2020oscar} and UNITER~\cite{chen2020uniter}. These use an object detector to obtain image features, then concatenate image features and text features and input the concatenation into a Transformer~\cite{vaswani2017attention} module to learn the joint image-text embedding. None of these methods can conduct image-text alignment before fusion. This problem can cause difficulty in learning the interaction between images and text. To obtain better interaction, ALBEF~\cite{li2021align} introduces a contrastive loss to align image and text features before fusing them and then generates joint representations. In this work, we are interested in how to leverage a pre-trained vision-language model to further improve class incremental learning.

\section{Method}
\begin{figure*}[t]
\centerline{\includegraphics[width=0.85\textwidth]{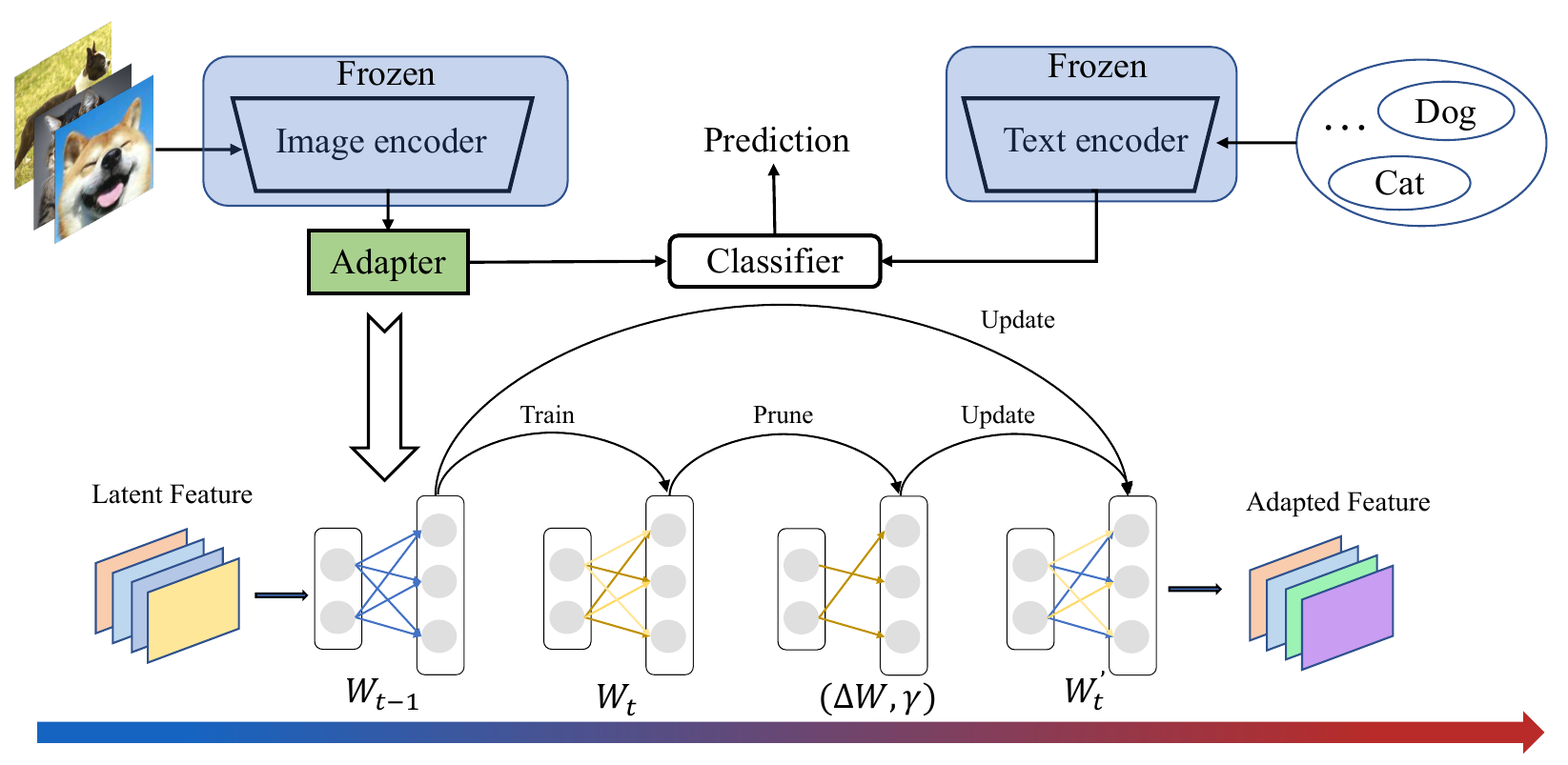}}
\vspace{-8pt}
\caption{Illustration of our proposed method. We add a Linear Adapter layer (in green) after the CLIP Image Encoder which generates projected features. The text feature is generated by encoding class labels using the CLIP Text Encoder. During training the Image and Text encoders are frozen and we only update the Linear Adapter. To obtain better continual learning performance, we propose a parameter retention method to mitigate forgetting in the Adapter (see Parameter Retention Section).}

\label{fig:idea}
\end{figure*}

In Figure~\ref{fig:idea}, we present the overall framework of the proposed method, encompassing the CLIP pre-trained image encoder and text encoder, as well as a linear adapter. During the training process, we solely update this linear adapter and keep all other parts fixed. Additionally, we introduce a parameter retention method that maximizes the preservation of knowledge acquired from previous tasks, thereby mitigating catastrophic forgetting. 
We begin by describing the CIL scenario we consider and then describe our approach.

\subsection{Preliminaries}

\minisection{Class incremental learning.}
In class incremental learning a model must be updated over $T$ sequentially-presented classification tasks. At task $t$ the data available for training is $\mathcal{D}_t = {(\mathbf{x}_t^{(i)}, y_t^{(i)})}_{i=1}^{n_t}$, where $\mathbf{x}_t^{(i)}$ is an input image, $y_t^{(i)}$ is the corresponding label, and $n_t$ is the number of samples in task $t$. We write the number of new classes at task $t$ as $K_t$ and $K_{1:t} = \sum_{t'=1}^t K_t$ for the total number of classes up to the current task. After learning on $\mathcal{D}_t$, the model is evaluated on all previous tasks -- that is, we require that the model continue to perform well on previously seen tasks. 

Normally the model can be divided into two parts: a feature extractor $F_\theta$ and a fully-connected classification head $H_\phi$ with parameters $\theta$ and $\phi$, respectively. A fully-connected layer normally consists of a weight matrix and a bias vector, which we denote as $\phi=\{\mathbf{W}, \mathbf{b}\}$. The conventional cross-entropy loss is often used for learning to classify, which for task $t$ is:
\begin{equation}
\label{eq:entropy}
    \mathcal{L}_{\text{CE}} (\mathbf{x}, \mathbf{y}; \theta, \phi) = -\frac{1}{n} \sum_{i=1} ^{n} \mathbf{y}_{i}\cdot \log \mathbf{\mathbf{p}}_{i} (\mathbf{x}),
\end{equation}
where $\mathbf{y}_{i}$ is the one-hot vector encoding the correct label and $\mathbf{p}_{i}$ are the model probabilities obtained by softmax over the outputs of the classifier head:
\begin{equation}
\label{eq:softmax}
    \mathbf{p}_{i}(\mathbf{x}) = \frac{\exp{(\mathbf{W}_{i}F(\mathbf{x})+b_{i})}}{\sum_{j=1}^{K_{1:t}}
    \exp{(\mathbf{W}_{j}F(\mathbf{x})+b_{j})}},
\end{equation}
where $K_{1:t}$ is the number of classes up the task $t$ and $\mathbf{W}_{i}$ is the weight vector corresponding to $i_{\text{th}}$ class.

The challenge of class incremental learning is to minimize the loss in Eq.~(\ref{eq:entropy}) using \emph{only} data from task $t$ without forgetting how to recognize the $K_{1:t-1}$ classes from all previous tasks.  

\minisection{CLIP for continual learning.}
\label{continual-clip}
Thengane et al.~\cite{continual-clip} proposed exploiting the zero-shot learning capability of CLIP to mitigate catastrophic forgetting of previous tasks during continual learning. Continual-CLIP~\cite{continual-clip} classifies images by multiplying the encoded image with text features extracted using the text encoder in a frozen CLIP model using the labels of all categories seen up to the current task as prompts. The prompts are manually determined and combined with the class name and the similarity score between the input image and different prompts from all classes is used for classification. This method does not require storing any exemplars or updating any parameters. It outperforms most state-of-the-art methods with only zero-shot evaluation. Continual-CLIP performs no adaptation ability on incoming task data and in the next section, we propose how CLIP pre-trained models can be updated and further improved for class incremental learning.

\subsection{CLIP with Adaptation}
\label{sec:adaptation}
Starting from the zero-shot evaluation protocol proposed in Continual-CLIP~\cite{continual-clip}, we aim to augment the original architecture with new modules that enable better adaptation to more downstream tasks. By learning on the training set of each task and updating the parameters of these modules, we can achieve a better trade-off between stability and plasticity.

\minisection{Adaptation via a Linear Adapter.}
Assume we are at task $t$, the input image is $x_{t}^{(i)}$, and denote the CLIP image encoder as $F_{\text{image}}$ and the text encoder as $F_{\text{text}}$. For the image modality, we input the images into the image encoder to extract the latent embedding $I_{i}=F_{\text{image}}(x_{t}^{i})\in \mathbb{R}^{M} $. Subsequently, 

we pass it through a Linear Adapter layer $A_{i} = g_W(I_{i})$ parameterized by weight matrix $W\in \mathbb{R}^{M\times M}$ to compute an adapted image feature with more capacity. For the text modality, we amalgamate a manually specified, fixed set of prompts (``a photo of [CLS]'') where [CLS] takes the value of all class names $C_t=\{c_1, c_2, ..., c_t\}$ that the current model has encountered. This set of prompts is given to the text encoder to obtain the text features $B_t$. Finally, we perform a matrix multiplication between $A_i$ and $B_{t}$ to compute the output logits. We use the cross-entropy loss to update the parameters of the Linear Adapter layer:
\begin{equation}
\label{eq:entropy-our}
    \mathcal{L}_{\text{CE}} (\mathbf{x}, y, t; W) = -\frac{1}{n} \sum_{i=1} ^{n} {y}_{i}\cdot \log {A_{i}}{B_t}.
\end{equation}

\minisection{Adaptation via a Self-attention Adapter.}
\label{self-att}
We can use a self-attention module in place of the single Linear Adapter layer described above. Specifically, we apply three separate linear transformations to the output $I_i$ of the CLIP image encoder, resulting in the query ($Q$), key ($K$), and value ($V$) matrices:
\begin{equation}
\label{eq:attention}
    Q = W_q  I_i,
    K = W_k  I_i,
    V = W_v  I_i,
\end{equation}
where $W_q$, $W_k$, and $W_v$ are learned weight matrices. Next, we compute the self-attention weights using the dot product between the query and key matrices, scaled by a factor $\sqrt{M}$:
\begin{equation}
    \alpha = \text{softmax}(\frac{QK^T}{\sqrt{M}} ),
\end{equation}
Finally, we compute the weighted sum of the value matrix V, using the self-attention weights $\alpha$:
\begin{equation}
    A_i = \alpha V.
\end{equation}

After obtaining the adapted image features $A_{i}$, we use Eq.~\ref{eq:entropy-our} to calculate the logits by multiplying the image features with the text features and use the cross-entropy loss to update parameters. Unlike the previous approach, in this Self-attention Adapter we need to update three linear projection matrices instead of just one. This results in three times the number of learnable parameters.

\minisection{Adaptation via Prompt Tuning.}
The two methods above operate only on the image modality. We propose adding learnable parameters to the text prompt. In the previous two methods, we use fixed prompts such as ``a photo of [CLS]''. For prompt tuning, we follow the CoOp method~\cite{CoOp} and replace the fixed prompt with a learnable parameter $p$. We concatenate it with the first layer of embedded input tokens for the class prompts that the model has seen so far, and then give it in input to $F_{\text{text}}$ to obtain the text features $B_t$.  In this method, we use the fixed CLIP image encoder-generated $I_i$ directly as $A_i$ and multiply it with $B_t$. We then calculate the loss in Eq.~\ref{eq:entropy-our} and update the prompt vector to optimize the model parameters.

\subsection{Parameter Retention}
\label{sec:retention}
Although a single adapter layer may achieve satisfactory performance on the current task, it will drift in the learning process and result in catastrophic forgetting in the later stages of continual learning, leading to a decline in overall performance. Therefore, we propose a parameter retention strategy aimed at maximizing the preservation of previously learned knowledge while learning new tasks.

Using a simple Linear Adapter as an example, before the start of training we randomly initialize an $M\times M$ parameter matrix $W_0$ for the Linear Adapter. After completing training and testing the first task, we save the trained parameter matrix $W_1$ and use it to initialize the adapter for the second task. After training the second task, $W_2$ is selectively updated to preserve a proportion of the original parameters from $W_1$. Measuring the drift of parameter $i$ in $W$ with $\Delta W^i = |W^i_1-W^i_2|$, we select a proportion $\gamma \in [0, 1.0]$ of parameters from $W_1$, while the rest are replaced with the new parameters from $W_2$:
\begin{equation}
    W_2^{\prime,i} =\left\{\begin{matrix}
W_1^i, \quad \text{if}  \ \Delta W^i < \beta(\gamma),
\\ 
W_2^i, \quad \text{if} \  \Delta W^i \geq \beta(\gamma),
\end{matrix}\right.
\end{equation}
where $\beta(\gamma)$ is a dynamic threshold computed from the desired retention rate $\gamma$ and a ranking of the parameter drifts $\Delta W^i$ (i.e. we order the values in $\Delta W^i$ and select the threshold $\beta$ that preserves exactly the desired proportion $\gamma$ of weights).
We use the new parameter matrix ${W_2}^{\prime}$ to test the model. Before the end of each subsequent task, we use the same method to preserve the parameters at all stages of continual learning.

This parameter retention strategy is inspired by the human cognitive process. There are studies indicating that humans activate different neurons in the brain when acquiring different forms of knowledge~\cite{pulvermuller2005brain}. By modifying only a portion of the neural network weights while retaining most of the original parameters during the learning of new categories, our model can, to a certain extent, emulate this process and explicitly retain prior knowledge it has acquired.

As for the question of \textit{which} data to retain, we believe that preserving the parts with the most significant changes is sufficient for the model to make adequate adjustments to adapt to new tasks. We also conducted ablation experiments using other methods, such as randomly preserving parameters, and the results confirmed the validity of our approach (see Ablation Section).

\minisection{Knowledge distillation.}
Distillation is widely employed in continual learning. The outputs of a previous model are used to impose certain constraints on the parameters of the new one. We attempted to incorporate knowledge distillation in our framework by distilling the output logits from both the current and previous models. However, the results using distillation achieve worse performance (see Experimental Section).

\begin{figure*}
\centerline{\includegraphics[width=0.90\textwidth]{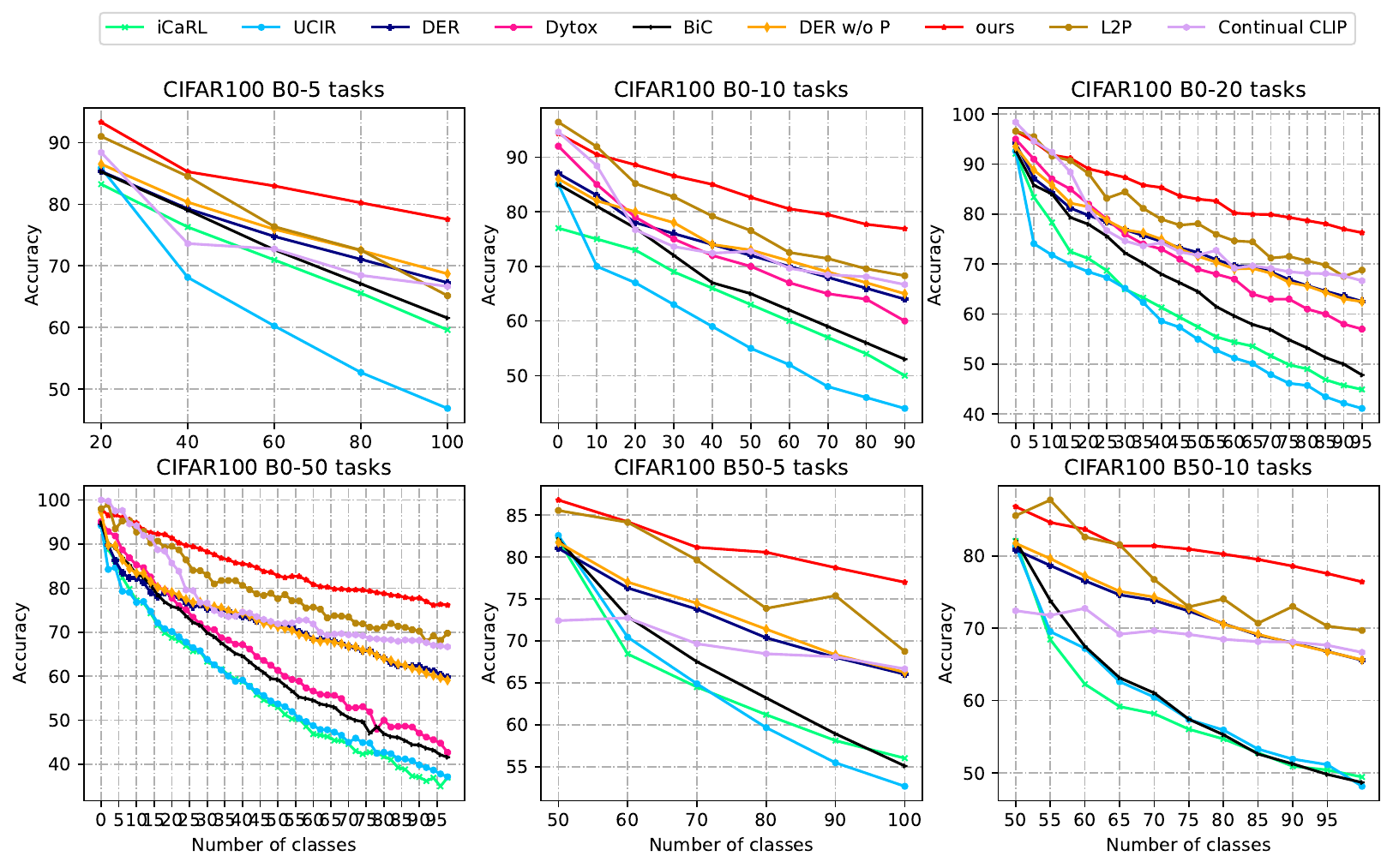}}
\vspace{-8pt}
\caption{Comparison of our method with the state-of-the-art on CIFAR-100 in different experimental settings.
}

\label{fig:overview}
\end{figure*}

\section{Experimental Results}
We first discuss implementation details, benchmarks, and competing methods. Then we provide a  comparison to the state-of-the-art and an extensive ablation study.

\subsection{Experimental Setup}
We performed experiments on CIFAR-100, ImageNet-100 (also known as ImageNet-Subset) \textcolor{black}{and ImageNet-R~\cite{hendrycks2021many}}. First, we follow the two experimental configurations used in DER~\cite{yan2021dynamically}: B0, in which all classes are equally divided among different tasks, and B50 in which the first task contains 50 classes (half of the dataset) and the rest are equally divided into subsequent tasks. \textcolor{black}{Then we follow the ImageNet-R setting proposed by L2P~\cite{l2p} which splits the 200 classes into 10 tasks equally.} For CIFAR-100, we train using SGD for 30 epochs at an initial learning rate of 0.1 and weight decay of 0.0002. For ImageNet-100 \textcolor{black}{and ImageNet-R}, we use Adam with an initial learning rate of 0.01 for 10 epochs. For both datasets we use a cosine annealing scheduler. We report \emph{average incremental accuracy (Avg)} as a function of increasing tasks and the accuracy at the last task (Last) when one number is preferable. All results are averaged over 3 runs with different class orders and random seeds.

We compare our approach with several state-of-the-art methods: iCaRL~\cite{rebuffi2017icarl}, WA~\cite{zhao2020maintaining} UCIR~\cite{hou2019learning}, PODNet~\cite{douillard2020podnet}, BiC~\cite{wu2019large}, RPSNet~\cite{rajasegaran2019random}, and DER~\cite{yan2021dynamically}. Unless otherwise stated we store a total of 2000 exemplars for all approaches except Continual-CLIP (which is based on zero-shot inference using the pre-trained CLIP model and does no adaptation to incremental tasks).

Many of the state-of-the-art methods mentioned above are based on ResNet backbones and do not use pre-trained models. Therefore, we have also developed some baselines using pre-trained models.
The \textbf{Linear Probe} method adds a classifier on top of a fixed CLIP image encoder. As new tasks arrive, the classifier head is incrementally expanded, and its parameters are retrained to classify the new categories.
The \textbf{Continual-CLIP} method~\cite{continual-clip} relies solely on the excellent generalization capabilities of CLIP itself, without adding any additional parameters or further training. 

The \textbf{L2P}~\cite{l2p} method uses a dynamic prompt selection approach as input to a pre-trained Transformer. Unlike the original implementation that uses ImageNet-21K pre-trained weights as the backbone, we replaced it with the CLIP image encoder since CIFAR-100 and ImageNet-100 have overlapping classes with 21K classes, which would bias the results.

\subsection{Comparison with the State-of-the-art}

\begin{table*}
\begin{center}
\resizebox{0.8\textwidth}{!}{%
\begin{tabular}{r|cccccc|cccc|c}
\toprule
\multirow{4}{*}{Method} & \multicolumn{6}{c|}{ImageNet-100 B0}                                                                                                               & \multicolumn{4}{c|}{ImageNet-100 B50}                                                                       & ImageNet-R B0 \\ \cline{2-12} 
                        & \multicolumn{6}{c|}{\# tasks}                                                                                                                      & \multicolumn{4}{c|}{\# tasks}                                                                               & tasks         \\
                        & \multicolumn{2}{c|}{5}                               & \multicolumn{2}{c|}{10}                              & \multicolumn{2}{c|}{20}              & \multicolumn{2}{c|}{5}                               & \multicolumn{2}{c|}{10}                              & 10            \\ \cline{2-12} 
                        & \multicolumn{1}{c|}{Avg} & \multicolumn{1}{c|}{Last} & \multicolumn{1}{c|}{Avg} & \multicolumn{1}{c|}{Last} & \multicolumn{1}{c|}{Avg} & Last      & \multicolumn{1}{l|}{Avg} & \multicolumn{1}{l|}{Last} & \multicolumn{1}{l|}{Avg} & \multicolumn{1}{l|}{Last} & Last          \\ \hline
iCaRL                   & 78.1                     & \multicolumn{1}{c|}{65.2} & 74.1                     & \multicolumn{1}{c|}{58.5} & 69.0                     & 50.9      & 60.7                     & \multicolumn{1}{c|}{44.7} & 57.3                     & 44.4                      & -             \\
End2End                 & 75.5                     & \multicolumn{1}{c|}{64.0} & 70.1                     & \multicolumn{1}{c|}{50.3} & 68.3                     & 48.9      & 61.0                     & \multicolumn{1}{c|}{52.1} & 58.5                     & 52.2                      & -             \\
UCIR                   & 76.0                     & \multicolumn{1}{c|}{64.0} & 70.5                     & \multicolumn{1}{c|}{55.3} & 64.7                     & 47.8      & 77.2                     & \multicolumn{1}{c|}{68.2} & 66.9                     & 56.8                      & -             \\
PODNet                  & 78.2                     & \multicolumn{1}{c|}{66.2} & 72.3                     & \multicolumn{1}{c|}{57.2} & 66.7                     & 48.9      & 80.3                     & \multicolumn{1}{c|}{73.5} & 79.0                     & 70.8                      & -             \\
UCIR-DDE               & 77.2                     & \multicolumn{1}{c|}{65.8} & 71.7                     & \multicolumn{1}{c|}{56.8} & 66.2                     & 40.0      & 78.8                     & \multicolumn{1}{c|}{68.1} & 68.4                     & 57.9                      & -             \\
RM                      & 75.5                     & \multicolumn{1}{c|}{62.2} & 70.4                     & \multicolumn{1}{c|}{53.2} & 65.4                     & 45.7      & 56.9                     & \multicolumn{1}{c|}{41.8} & 57.7                     & 37.3                      & -             \\
DER                     & -                        & \multicolumn{1}{c|}{-}    & 77.2                     & \multicolumn{1}{c|}{66.7} & -                        & -         & -                        & \multicolumn{1}{c|}{-}    & 78.2                     & 74.9                      & 66.7          \\ \midrule

Continual-CLIP          & 81.7                     & \multicolumn{1}{c|}{75.3} & 83.1                     & \multicolumn{1}{c|}{75.3} & 83.5                     & 75.3      & 79.2                     & \multicolumn{1}{c|}{75.3} & 79.3                     & 75.3                      & 72.0          \\ 
Linear Probe            & 79.4                     & \multicolumn{1}{c|}{64.5} & 81.8                     & \multicolumn{1}{c|}{67.4} & 84.0                     & 72.0      & 81.4                     & \multicolumn{1}{c|}{67.2} & 83.1                     & 72.0                      & 57.5          \\
L2P                     & 81.3                     & \multicolumn{1}{c|}{\bf{76.3}} & 80.1                     & \multicolumn{1}{c|}{75.8} & 80.7                     & 76.1      & 74.8                     & \multicolumn{1}{c|}{74.1} & 71.9                     & 72.5                      & 74.1          \\ \midrule
Ours                    & \bf{84.9}                & \multicolumn{1}{c|}{74.7} & \bf{85.3}                & \multicolumn{1}{c|}{\bf{76.8}} & \bf{84.1}                & \bf{77.0} & \bf{85.1}                & \multicolumn{1}{c|}{\bf{76.8}} & \bf{85.4}                & \bf{77.4}                 & \bf{75.9}     \\ \bottomrule
\end{tabular}%
}
\end{center}
\vspace{-8pt}
\caption{Comparison of our method with the state-of-the-art on ImageNet-100 under different experimental settings.}
\label{tab:imagenet}
\label{tab:my-table}
\end{table*}

\minisection{Evaluation on CIFAR-100.} 
In Figure~\ref{fig:overview} we plot the performance of our model on each task under different settings. We see that our model outperforms others in almost all settings and for every task. This shows that we not only exploit the performance brought by the large-scale pre-trained CLIP model, but that it is also adapting to the continual learning scenario. It is worth noting that in the extremely challenging B0-50 setting, in which there are 50 continual learning tasks, our method does not show a clear advantage over other state-of-the-art models in the very early tasks. This is due to the fact that, in the early stages of continual learning, we directly replace a selected portion of parameters with the previous ones, which can cause some degree of underfitting in tasks with fewer classes. However, in later tasks our method can preserve knowledge from previous tasks much better, which mitigates forgetting and ultimately leads to superior performance compared to other methods. We refer the readers to the supplementary material for more detailed results.

\minisection{Evaluation on ImageNet-100.}
Like the experiments on CIFAR-100, we follow DER~\cite{yan2021dynamically} and evaluate on ImageNet-100 with the same settings to compare with existing methods. As shown in Table~\ref{tab:imagenet}, the current state-of-the-art method DER outperforms iCaRL, End2End, and UCIR by a large margin. Our method improves over DER by a significant gain of up to 8\% points in the B0-10 setting and 7\% points in the B50-10 setting in the Avg metric. In general, our method yields larger gains in the B0 setting than in B50 compared to DER. 
It's worth noting that when employing CLIP pre-trained weights and utilizing an exemplar set of 2000, the improvement in L2P's performance over Continual-CLIP is negligible. In most settings, L2P's performance is lower than ours about 1\% to 2\%.

\minisection{Evaluation on ImageNet-R.}
We follow ~\cite{hu2023pop} using an exemplar of size 5000 for ImageNet-R.
In Table~\ref{tab:imagenet}, we see that our results surpass DER by 9\%. While L2P outperforms Continual-CLIP by 2.1\%, there still remains a gap of 1.9\% compared to our results. This further demonstrates the superiority of our method in effectively utilizing pre-trained models to enhance the performance of continual learning.

\subsection{Ablation Study}
\label{sec:ablation}
We conducted an ablation study in the B0-10 setting on CIFAR-100. First, we evaluate different adapters, and then we look at the effects of parameter retention rates and other strategies for parameter retention. Finally, we analyze how the number of exemplars affects performance.

\begin{table}
\centering
\resizebox{.48\textwidth}{!}
{%
\begin{tblr}{
  vline{2} = {-}{},
  hline{1-2,4} = {-}{},
  cells = {c}
}
Adapter & Continual-CLIP & Self-Attention & Prompt Tuning  &MLP&  Linear (Ours)   \\
Avg      & 75.1     & 75.8    & 76.4 &81.9& 84.2\\
Last     & 66.7     & 71.4    & 59.0 &70.8& 76.9 
\end{tblr}
}
\vspace{-6pt}
\caption{Our performance with different adapters in average and last-task accuracy in the CIFAR-100 B0-10 setting.}
\label{ab:adaptor}
\end{table}

\minisection{Different adapter layers.}
As shown in Table~\ref{ab:adaptor}, Continual-CLIP leverages the zero-shot capabilities of CLIP and can exploit the pre-trained CLIP model without any further learning. The Linear Adapter is the best architecture we found for continual learning, and it requires only one additional fully-connected layer. We also evaluated more fully-connected layers (MLP), but it does not perform well. Self-attention can leverage more parameters than the Linear Adapter method, but does not improve the results. This may be due to the complexity of the self-attention architecture. Prompt Tuning learns additional text prompts, however in the continual learning setting these learned prompts can drift away, resulting in much worse results.

\begin{table}
\centering
\resizebox{.4\textwidth}{!}{%
\begin{tblr}{
  vline{2} = {-}{},
  hline{1-2,4} = {-}{},
  cells = {c}
}
Ratio ($\gamma$) & 0              & 0.2 & 0.4 & 0.6 & 0.8 &0.9 \\
Avg   & 83.2     & 83.3 & 83.5 & 84.0 & 84.2 &81.6  \\
Last  & 73.2     & 73.5 & 74.1 & 74.7 & 76.9 &76.5   
\end{tblr}
}
\vspace{-6pt}
\caption{Our performance with different retention ratios on the CIFAR-100 B0-10 setting. The ratio is the percentage of parameters retained from the previous adapter.}
\label{ab:ratio}
\end{table}

\minisection{Different parameter retention ratios.}
In Table~\ref{ab:ratio} we show the impact of parameter retention rates on performance. Note that $\gamma = 0$ means we retain no parameters and directly use the adapted ones for testing after training a new task. We see that retaining more parameters leads to better performance, but retaining more than $\gamma = 0.8$ causes the model to underfit to new tasks and leads to a decrease in performance. We use $\gamma = 0.8$ for all our experiments.

\begin{table}
\centering
\resizebox{.27\textwidth}{!}{%
\begin{tblr}{
  cell{2}{2} = {c},
  cell{2}{3} = {c},
  cell{2}{4} = {c},
  cell{3}{2} = {c},
  cell{3}{3} = {c},
  cell{3}{4} = {c},
  vline{2} = {-}{},
  hline{1-2,4} = {-}{},
  cells = {c}
}
Method  & KD   & Random & Ours \\
Avg     & 76.7 & 44.5   & 84.2 \\
Last    & 67.0 & 42.2   & 76.9 
\end{tblr}
}
\vspace{-6pt}
\caption{Our performance using different parameter retention strategies in the CIFAR-100
B0-10 setting.}
\label{ab:kd}
\vspace{-6pt}
\end{table}

\minisection{Comparison of parameter retention strategies.}
In Table~\ref{ab:kd} we compare our parameter retention strategy with Knowledge Distillation (KD) and Random Selection. KD distills the logits from the previous model to regularize model adaptation to new tasks, however it results in worse performance compared to the original method without it. Distillation may be useful in some models trained from scratch, since there are more adjustable parameters and more precise optimizations can be made. However, in our model with only an additional linear layer, complex distillation is less effective than simply retaining some useful parameters. As expected, if we retain parameters randomly instead of using our selection method, CIL completely fails since important parameters are not explicitly retained to balance stability and plasticity.

\begin{table}
\centering
\resizebox{.42\textwidth}{!}{%
\begin{tblr}{
  cell{2}{2} = {c},
  cell{2}{3} = {c},
  cell{2}{4} = {c},
  cell{2}{5} = {c},
  cell{3}{2} = {c},
  cell{3}{3} = {c},
  cell{3}{4} = {c},
  cell{3}{5} = {c},
  vline{2} = {-}{},
  hline{1-2,4} = {-}{},
  cells = {c}
}
\# Exemplars & 2000 & 1000 & 500  &  Linear Probe (2000) \\
Avg       & 84.2 & 82.5 & 80.0 & 80.5               \\
Last      & 76.9 & 71.9 & 68.2 & 66.8               
\end{tblr}
}
\vspace{-6pt}
\caption{Our performance for different numbers of exemplars in the CIFAR-100 B0-10 setting.}
\label{tab:ab-exp}
\end{table}

\minisection{Ablation on the number of exemplars.}
Finally, we show the impact of the number of exemplars in Table~\ref{tab:ab-exp}. As expected, the more exemplars used, the better the performance. Reducing from 2000 to 500 exemplars degrades performance. It is notable that by using 500 exemplars, our method can achieve performance similar to Linear Probe with 2000, demonstrating the effectiveness of our proposed method.

\section{Conclusions}

In this work we approached the problem of class incremental learning using a pre-trained vision-language model. We investigated three different adaptation strategies (a Linear Adapter, a Self-attention Adapter, and Prompt Tuning) to facilitate model adaptation for class incremental learning. We found that using a Linear Adapter is simple but effective compared to the other two methods. Adapting the CLIP Image Encoder alone yields significant performance gains over conventional methods and baselines with pre-trained models on all benchmarks and in all incremental learning scenarios. For future work we are exploring more effective ways of leveraging pre-trained models for incremental learning in order to take full advantage of the rapid development of multi-modal foundation models.

\bigskip

\bibliography{paper}

\begin{thebibliography}{57}
\providecommand{\natexlab}[1]{#1}

\bibitem[{Ahn et~al.(2021)Ahn, Kwak, Lim, Bang, Kim, and Moon}]{ahn2021ss}
Ahn, H.; Kwak, J.; Lim, S.; Bang, H.; Kim, H.; and Moon, T. 2021.
\newblock SS-IL: Separated Softmax for Incremental Learning.
\newblock In \emph{ICCV}.

\bibitem[{Aljundi et~al.(2018)Aljundi, Babiloni, Elhoseiny, Rohrbach, and Tuytelaars}]{aljundi2018memory}
Aljundi, R.; Babiloni, F.; Elhoseiny, M.; Rohrbach, M.; and Tuytelaars, T. 2018.
\newblock Memory aware synapses: Learning what (not) to forget.
\newblock In \emph{ECCV}.

\bibitem[{Bang et~al.(2021)Bang, Kim, Yoo, Ha, and Choi}]{bang2021rainbow}
Bang, J.; Kim, H.; Yoo, Y.; Ha, J.-W.; and Choi, J. 2021.
\newblock Rainbow memory: Continual learning with a memory of diverse samples.
\newblock In \emph{CVPR}.

\bibitem[{Belouadah and Popescu(2019)}]{belouadah2019il2m}
Belouadah, E.; and Popescu, A. 2019.
\newblock Il2m: Class incremental learning with dual memory.
\newblock In \emph{ICCV}.

\bibitem[{Belouadah, Popescu, and Kanellos(2020)}]{belouadah2020comprehensive}
Belouadah, E.; Popescu, A.; and Kanellos, I. 2020.
\newblock A comprehensive study of class incremental learning algorithms for visual tasks.
\newblock \emph{Neural Networks}.

\bibitem[{Castro et~al.(2018)Castro, Mar{\'\i}n-Jim{\'e}nez, Guil, Schmid, and Alahari}]{castro2018end}
Castro, F.~M.; Mar{\'\i}n-Jim{\'e}nez, M.~J.; Guil, N.; Schmid, C.; and Alahari, K. 2018.
\newblock End-to-end incremental learning.
\newblock In \emph{ECCV}.

\bibitem[{Chen et~al.(2020)Chen, Li, Yu, El~Kholy, Ahmed, Gan, Cheng, and Liu}]{chen2020uniter}
Chen, Y.-C.; Li, L.; Yu, L.; El~Kholy, A.; Ahmed, F.; Gan, Z.; Cheng, Y.; and Liu, J. 2020.
\newblock Uniter: Universal image-text representation learning.
\newblock In \emph{ECCV}.

\bibitem[{Cossu et~al.(2022)Cossu, Tuytelaars, Carta, Passaro, Lomonaco, and Bacciu}]{cossu2022continual}
Cossu, A.; Tuytelaars, T.; Carta, A.; Passaro, L.; Lomonaco, V.; and Bacciu, D. 2022.
\newblock Continual pre-training mitigates forgetting in language and vision.
\newblock \emph{arXiv preprint arXiv:2205.09357}.

\bibitem[{Delange et~al.(2021)Delange, Aljundi, Masana, Parisot, Jia, Leonardis, Slabaugh, and Tuytelaars}]{delange2021continual}
Delange, M.; Aljundi, R.; Masana, M.; Parisot, S.; Jia, X.; Leonardis, A.; Slabaugh, G.; and Tuytelaars, T. 2021.
\newblock A continual learning survey: Defying forgetting in classification tasks.
\newblock \emph{TPAMI}.

\bibitem[{Dosovitskiy et~al.(2020)Dosovitskiy, Beyer, Kolesnikov, Weissenborn, Zhai, Unterthiner, Dehghani, Minderer, Heigold, Gelly et~al.}]{dosovitskiy2020image}
Dosovitskiy, A.; Beyer, L.; Kolesnikov, A.; Weissenborn, D.; Zhai, X.; Unterthiner, T.; Dehghani, M.; Minderer, M.; Heigold, G.; Gelly, S.; et~al. 2020.
\newblock An image is worth 16x16 words: Transformers for image recognition at scale.
\newblock \emph{ICLR}.

\bibitem[{Douillard et~al.(2020)Douillard, Cord, Ollion, Robert, and Valle}]{douillard2020podnet}
Douillard, A.; Cord, M.; Ollion, C.; Robert, T.; and Valle, E. 2020.
\newblock Podnet: Pooled outputs distillation for small-tasks incremental learning.
\newblock In \emph{ECCV}.

\bibitem[{Ermis et~al.(2022)Ermis, Zappella, Wistuba, and Archambeau}]{ermis2022memory}
Ermis, B.; Zappella, G.; Wistuba, M.; and Archambeau, C. 2022.
\newblock Memory efficient continual learning with transformers.
\newblock \emph{NIPS}.

\bibitem[{Hayes et~al.(2020)Hayes, Kafle, Shrestha, Acharya, and Kanan}]{hayes2020remind}
Hayes, T.~L.; Kafle, K.; Shrestha, R.; Acharya, M.; and Kanan, C. 2020.
\newblock Remind your neural network to prevent catastrophic forgetting.
\newblock In \emph{ECCV}.

\bibitem[{Hendrycks et~al.(2021)Hendrycks, Basart, Mu, Kadavath, Wang, Dorundo, Desai, Zhu, Parajuli, Guo, Song, Steinhardt, and Gilmer}]{hendrycks2021many}
Hendrycks, D.; Basart, S.; Mu, N.; Kadavath, S.; Wang, F.; Dorundo, E.; Desai, R.; Zhu, T.; Parajuli, S.; Guo, M.; Song, D.; Steinhardt, J.; and Gilmer, J. 2021.
\newblock The Many Faces of Robustness: A Critical Analysis of Out-of-Distribution Generalization.
\newblock \emph{ICCV}.

\bibitem[{Hou et~al.(2019)Hou, Pan, Loy, Wang, and Lin}]{hou2019learning}
Hou, S.; Pan, X.; Loy, C.~C.; Wang, Z.; and Lin, D. 2019.
\newblock Learning a Unified Classifier Incrementally via Rebalancing.
\newblock In \emph{ICCV}.

\bibitem[{Hu et~al.(2021)Hu, Tang, Miao, Hua, and Zhang}]{hu2021distilling}
Hu, X.; Tang, K.; Miao, C.; Hua, X.-S.; and Zhang, H. 2021.
\newblock Distilling causal effect of data in class-incremental learning.
\newblock In \emph{CVPR}.

\bibitem[{Hu et~al.(2023)Hu, Lyu, Gao, and Vasconcelos}]{hu2023pop}
Hu, Z.; Lyu, J.; Gao, D.; and Vasconcelos, N. 2023.
\newblock POP: Prompt Of Prompts for Continual Learning.
\newblock arXiv:2306.08200.

\bibitem[{Janson et~al.(2023)Janson, Zhang, Aljundi, and Elhoseiny}]{janson2022simple}
Janson, P.; Zhang, W.; Aljundi, R.; and Elhoseiny, M. 2023.
\newblock A Simple Baseline that Questions the Use of Pretrained-Models in Continual Learning.
\newblock arXiv:2210.04428.

\bibitem[{Jia et~al.(2021)Jia, Yang, Xia, Chen, Parekh, Pham, Le, Sung, Li, and Duerig}]{jia2021scaling}
Jia, C.; Yang, Y.; Xia, Y.; Chen, Y.-T.; Parekh, Z.; Pham, H.; Le, Q.; Sung, Y.-H.; Li, Z.; and Duerig, T. 2021.
\newblock Scaling up visual and vision-language representation learning with noisy text supervision.
\newblock In \emph{ICML}.

\bibitem[{Kirkpatrick et~al.(2017)Kirkpatrick, Pascanu, Rabinowitz, Veness, Desjardins, Rusu, Milan, Quan, Ramalho, Grabska-Barwinska et~al.}]{kirkpatrick2017overcoming}
Kirkpatrick, J.; Pascanu, R.; Rabinowitz, N.; Veness, J.; Desjardins, G.; Rusu, A.~A.; Milan, K.; Quan, J.; Ramalho, T.; Grabska-Barwinska, A.; et~al. 2017.
\newblock Overcoming catastrophic forgetting in neural networks.
\newblock \emph{PNAS}.

\bibitem[{Li et~al.(2021)Li, Selvaraju, Gotmare, Joty, Xiong, and Hoi}]{li2021align}
Li, J.; Selvaraju, R.; Gotmare, A.; Joty, S.; Xiong, C.; and Hoi, S. C.~H. 2021.
\newblock Align before fuse: Vision and language representation learning with momentum distillation.
\newblock \emph{NIPS}.

\bibitem[{Li et~al.(2020)Li, Yin, Li, Zhang, Hu, Zhang, Wang, Hu, Dong, Wei et~al.}]{li2020oscar}
Li, X.; Yin, X.; Li, C.; Zhang, P.; Hu, X.; Zhang, L.; Wang, L.; Hu, H.; Dong, L.; Wei, F.; et~al. 2020.
\newblock Oscar: Object-semantics aligned pre-training for vision-language tasks.
\newblock In \emph{ECCV}.

\bibitem[{Liu et~al.(2018)Liu, Masana, Herranz, Van~de Weijer, Lopez, and Bagdanov}]{liu2018rotate}
Liu, X.; Masana, M.; Herranz, L.; Van~de Weijer, J.; Lopez, A.~M.; and Bagdanov, A.~D. 2018.
\newblock Rotate your networks: Better weight consolidation and less catastrophic forgetting.
\newblock In \emph{ICPR}.

\bibitem[{Liu, Schiele, and Sun(2021)}]{liu2021adaptive}
Liu, Y.; Schiele, B.; and Sun, Q. 2021.
\newblock Adaptive aggregation networks for class-incremental learning.
\newblock In \emph{CVPR}.

\bibitem[{Mallya, Davis, and Lazebnik(2018)}]{mallya2018piggyback}
Mallya, A.; Davis, D.; and Lazebnik, S. 2018.
\newblock Piggyback: Adapting a single network to multiple tasks by learning to mask weights.
\newblock In \emph{ECCV}.

\bibitem[{Mallya and Lazebnik(2018)}]{mallya2018packnet}
Mallya, A.; and Lazebnik, S. 2018.
\newblock Packnet: Adding multiple tasks to a single network by iterative pruning.
\newblock In \emph{CVPR}.

\bibitem[{Masana et~al.(2022)Masana, Liu, Twardowski, Menta, Bagdanov, and van~de Weijer}]{masana2020class}
Masana, M.; Liu, X.; Twardowski, B.; Menta, M.; Bagdanov, A.~D.; and van~de Weijer, J. 2022.
\newblock Class-incremental learning: survey and performance evaluation.
\newblock \emph{TPAMI}.

\bibitem[{Masana, Tuytelaars, and van~de Weijer(2021)}]{masana2020ternary}
Masana, M.; Tuytelaars, T.; and van~de Weijer, J. 2021.
\newblock Ternary feature masks: continual learning without any forgetting.
\newblock \emph{CVPR Workshops}.

\bibitem[{McCloskey and Cohen(1989)}]{mccloskey1989catastrophic}
McCloskey, M.; and Cohen, N.~J. 1989.
\newblock Catastrophic interference in connectionist networks: The sequential learning problem.
\newblock In \emph{Psychology of learning and motivation}, volume~24, 109--165. Elsevier.

\bibitem[{Parisi et~al.(2019)Parisi, Kemker, Part, Kanan, and Wermter}]{parisi2019continual}
Parisi, G.~I.; Kemker, R.; Part, J.~L.; Kanan, C.; and Wermter, S. 2019.
\newblock Continual lifelong learning with neural networks: A review.
\newblock \emph{Neural Networks}.

\bibitem[{Prabhu, Torr, and Dokania(2020)}]{prabhu2020gdumb}
Prabhu, A.; Torr, P.~H.; and Dokania, P.~K. 2020.
\newblock Gdumb: A simple approach that questions our progress in continual learning.
\newblock In \emph{ECCV}.

\bibitem[{Pulverm{\"u}ller(2005)}]{pulvermuller2005brain}
Pulverm{\"u}ller, F. 2005.
\newblock Brain mechanisms linking language and action.
\newblock \emph{Nature reviews neuroscience}.

\bibitem[{Radford et~al.(2021)Radford, Kim, Hallacy, Ramesh, Goh, Agarwal, Sastry, Askell, Mishkin, Clark et~al.}]{clip}
Radford, A.; Kim, J.~W.; Hallacy, C.; Ramesh, A.; Goh, G.; Agarwal, S.; Sastry, G.; Askell, A.; Mishkin, P.; Clark, J.; et~al. 2021.
\newblock Learning transferable visual models from natural language supervision.
\newblock In \emph{ICML}.

\bibitem[{Rajasegaran et~al.(2019)Rajasegaran, Hayat, Khan, Khan, and Shao}]{rajasegaran2019random}
Rajasegaran, J.; Hayat, M.; Khan, S.; Khan, F.~S.; and Shao, L. 2019.
\newblock Random path selection for incremental learning.
\newblock \emph{NIPS}.

\bibitem[{Razdaibiedina et~al.(2023)Razdaibiedina, Mao, Hou, Khabsa, Lewis, and Almahairi}]{razdaibiedina2023progressive}
Razdaibiedina, A.; Mao, Y.; Hou, R.; Khabsa, M.; Lewis, M.; and Almahairi, A. 2023.
\newblock Progressive Prompts: Continual Learning for Language Models.
\newblock arXiv:2301.12314.

\bibitem[{Rebuffi et~al.(2017)Rebuffi, Kolesnikov, Sperl, and Lampert}]{rebuffi2017icarl}
Rebuffi, S.-A.; Kolesnikov, A.; Sperl, G.; and Lampert, C.~H. 2017.
\newblock icarl: Incremental classifier and representation learning.
\newblock In \emph{CVPR}.

\bibitem[{Rusu et~al.(2022)Rusu, Rabinowitz, Desjardins, Soyer, Kirkpatrick, Kavukcuoglu, Pascanu, and Hadsell}]{rusu2016progressive}
Rusu, A.~A.; Rabinowitz, N.~C.; Desjardins, G.; Soyer, H.; Kirkpatrick, J.; Kavukcuoglu, K.; Pascanu, R.; and Hadsell, R. 2022.
\newblock Progressive Neural Networks.
\newblock arXiv:1606.04671.

\bibitem[{Schwarz et~al.(2018)Schwarz, Czarnecki, Luketina, Grabska-Barwinska, Whye~Teh, Pascanu, and Hadsell}]{schwarz2018progress}
Schwarz, J.; Czarnecki, W.; Luketina, J.; Grabska-Barwinska, A.; Whye~Teh, Y.; Pascanu, R.; and Hadsell, R. 2018.
\newblock Progress \& Compress: A scalable framework for continual learning.
\newblock In \emph{ICML}.

\bibitem[{Serra et~al.(2018)Serra, Suris, Miron, and Karatzoglou}]{serra2018overcoming}
Serra, J.; Suris, D.; Miron, M.; and Karatzoglou, A. 2018.
\newblock Overcoming Catastrophic Forgetting with Hard Attention to the Task.
\newblock In \emph{ICML}.

\bibitem[{Shin et~al.(2017)Shin, Lee, Kim, and Kim}]{shin2017continual}
Shin, H.; Lee, J.~K.; Kim, J.; and Kim, J. 2017.
\newblock Continual Learning with Deep Generative Replay.
\newblock \emph{NIPS}.

\bibitem[{Simon, Koniusz, and Harandi(2021)}]{simon2021learning}
Simon, C.; Koniusz, P.; and Harandi, M. 2021.
\newblock On learning the geodesic path for incremental learning.
\newblock In \emph{CVPR}.

\bibitem[{Tao et~al.(2020)Tao, Chang, Hong, Wei, and Gong}]{tao2020topology}
Tao, X.; Chang, X.; Hong, X.; Wei, X.; and Gong, Y. 2020.
\newblock Topology-preserving class-incremental learning.
\newblock In \emph{ECCV}.

\bibitem[{Thengane et~al.(2022)Thengane, Khan, Hayat, and Khan}]{continual-clip}
Thengane, V.; Khan, S.; Hayat, M.; and Khan, F. 2022.
\newblock CLIP model is an Efficient Continual Learner.
\newblock arXiv:2210.03114.

\bibitem[{Van~de Ven and Tolias(2019)}]{van2019three}
Van~de Ven, G.~M.; and Tolias, A.~S. 2019.
\newblock Three scenarios for continual learning.
\newblock \emph{NIPS Workshops}.

\bibitem[{Vaswani et~al.(2017)Vaswani, Shazeer, Parmar, Uszkoreit, Jones, Gomez, Kaiser, and Polosukhin}]{vaswani2017attention}
Vaswani, A.; Shazeer, N.; Parmar, N.; Uszkoreit, J.; Jones, L.; Gomez, A.~N.; Kaiser, {\L}.; and Polosukhin, I. 2017.
\newblock Attention is all you need.
\newblock \emph{NIPS}.

\bibitem[{Verwimp, De~Lange, and Tuytelaars(2021)}]{verwimp2021rehearsal}
Verwimp, E.; De~Lange, M.; and Tuytelaars, T. 2021.
\newblock Rehearsal revealed: The limits and merits of revisiting samples in continual learning.
\newblock In \emph{ICCV}.

\bibitem[{Wang et~al.(2022{\natexlab{a}})Wang, Zhang, Ebrahimi, Sun, Zhang, Lee, Ren, Su, Perot, Dy et~al.}]{wang2022dualprompt}
Wang, Z.; Zhang, Z.; Ebrahimi, S.; Sun, R.; Zhang, H.; Lee, C.-Y.; Ren, X.; Su, G.; Perot, V.; Dy, J.; et~al. 2022{\natexlab{a}}.
\newblock Dualprompt: Complementary prompting for rehearsal-free continual learning.
\newblock In \emph{ECCV}.

\bibitem[{Wang et~al.(2022{\natexlab{b}})Wang, Zhang, Lee, Zhang, Sun, Ren, Su, Perot, Dy, and Pfister}]{l2p}
Wang, Z.; Zhang, Z.; Lee, C.; Zhang, H.; Sun, R.; Ren, X.; Su, G.; Perot, V.; Dy, J.~G.; and Pfister, T. 2022{\natexlab{b}}.
\newblock Learning to Prompt for Continual Learning.
\newblock In \emph{CVPR}.

\bibitem[{Wu et~al.(2022)Wu, Swaminathan, Li, Ravichandran, Vasconcelos, Bhotika, and Soatto}]{wu2022class}
Wu, T.-Y.; Swaminathan, G.; Li, Z.; Ravichandran, A.; Vasconcelos, N.; Bhotika, R.; and Soatto, S. 2022.
\newblock Class-incremental learning with strong pre-trained models.
\newblock In \emph{CVPR}.

\bibitem[{Wu et~al.(2019)Wu, Chen, Wang, Ye, Liu, Guo, and Fu}]{wu2019large}
Wu, Y.; Chen, Y.; Wang, L.; Ye, Y.; Liu, Z.; Guo, Y.; and Fu, Y. 2019.
\newblock Large Scale Incremental Learning.
\newblock In \emph{ICCV}.

\bibitem[{Xue et~al.(2022)Xue, Zhang, Song, and Song}]{xue2022meta}
Xue, M.; Zhang, H.; Song, J.; and Song, M. 2022.
\newblock Meta-attention for ViT-backed Continual Learning.
\newblock In \emph{CVPR}.

\bibitem[{Yan, Xie, and He(2021)}]{yan2021dynamically}
Yan, S.; Xie, J.; and He, X. 2021.
\newblock DER: Dynamically Expandable Representation for Class Incremental Learning.
\newblock In \emph{CVPR}.

\bibitem[{Yu et~al.(2020)Yu, Twardowski, Liu, Herranz, Wang, Cheng, Jui, and Weijer}]{yu2020semantic}
Yu, L.; Twardowski, B.; Liu, X.; Herranz, L.; Wang, K.; Cheng, Y.; Jui, S.; and Weijer, J. v.~d. 2020.
\newblock Semantic drift compensation for class-incremental learning.
\newblock In \emph{CVPR}.

\bibitem[{Zenke, Poole, and Ganguli(2017)}]{zenke2017continual}
Zenke, F.; Poole, B.; and Ganguli, S. 2017.
\newblock Continual learning through synaptic intelligence.
\newblock In \emph{ICML}.

\bibitem[{Zhao et~al.(2020)Zhao, Xiao, Gan, Zhang, and Xia}]{zhao2020maintaining}
Zhao, B.; Xiao, X.; Gan, G.; Zhang, B.; and Xia, S.-T. 2020.
\newblock Maintaining discrimination and fairness in class incremental learning.
\newblock In \emph{CVPR}.

\bibitem[{Zhou et~al.(2022{\natexlab{a}})Zhou, Yang, Loy, and Liu}]{zhou2022conditional}
Zhou, K.; Yang, J.; Loy, C.~C.; and Liu, Z. 2022{\natexlab{a}}.
\newblock Conditional prompt learning for vision-language models.
\newblock In \emph{CVPR}.

\bibitem[{Zhou et~al.(2022{\natexlab{b}})Zhou, Yang, Loy, and Liu}]{CoOp}
Zhou, K.; Yang, J.; Loy, C.~C.; and Liu, Z. 2022{\natexlab{b}}.
\newblock Learning to Prompt for Vision-Language Models.
\newblock \emph{IJCV}.

\end{thebibliography}
\begin{table*}[]
\resizebox{\textwidth}{!}{%
\begin{tabular}{r|cccccccc}
\toprule
\multirow{4}{*}{Method}              & \multicolumn{8}{c}{CIFAR-100 B0}                                                                                                                                                                                                                                                 \\ \cline{2-9} 
                                     & \multicolumn{8}{c}{\# tasks}                                                                                                                                                                                                                                                     \\
                                     & \multicolumn{2}{c|}{5}                                             & \multicolumn{2}{c|}{10}                                            & \multicolumn{2}{c|}{20}                                            & \multicolumn{2}{c}{50}                                            \\ \cline{2-9} 
                                     & \multicolumn{1}{c|}{Avg} & \multicolumn{1}{c|}{Last}               & \multicolumn{1}{c|}{Avg} & \multicolumn{1}{c|}{Last}               & \multicolumn{1}{c|}{Avg} & \multicolumn{1}{c|}{Last}               & \multicolumn{1}{c|}{Avg} & Last                                   \\ \midrule
iCaRL         & 71.1                     & \multicolumn{1}{c|}{59.6}               & 65.3                     & \multicolumn{1}{c|}{50.9}               & 61.2                     & \multicolumn{1}{c|}{44.9}               & 56.1                     & 37.0                                   \\
LUCIR           & 62.8                     & \multicolumn{1}{c|}{46.9}               & 58.7                     & \multicolumn{1}{c|}{42.9}               & 58.2                     & \multicolumn{1}{c|}{41.1}               & 56.9                     & 37.2                                   \\
BiC                       & 73.1                     & \multicolumn{1}{c|}{61.5}               & 68.8                     & \multicolumn{1}{c|}{53.6}               & 66.5                     & \multicolumn{1}{c|}{47.8}               & 62.1                     & 41.6                                   \\
WA        & 72.8                     & \multicolumn{1}{c|}{60.3}               & 69.5                     & \multicolumn{1}{c|}{53.7}               & 67.3                     & \multicolumn{1}{c|}{48.2}               & 64.3                     & 42.7                                   \\
PODNet     & 66.7                     & \multicolumn{1}{c|}{51.5}               & 58.0                     & \multicolumn{1}{c|}{40.7}               & 54.0                     & \multicolumn{1}{c|}{35.8}               & 51.2                     & 33.4                                   \\
DER         & 76.8                     & \multicolumn{1}{c|}{67.3}               & 75.4                     & \multicolumn{1}{c|}{64.4}               & 74.1                     & \multicolumn{1}{c|}{62.6}               & 72.4                     & 59.8                                   \\
DyTox       & -                        & \multicolumn{1}{c|}{-}                  & 72.9                     & \multicolumn{1}{c|}{60.0}               & 72.2                     & \multicolumn{1}{c|}{57.0}               & 70.1                     & 52.0                                   \\ \midrule
L2P~                      & 77.9                     & \multicolumn{1}{c|}{65.2}               & 79.4                     & \multicolumn{1}{c|}{68.3}               & 79.5                     & \multicolumn{1}{c|}{68.8}               & 79.9                     & 69.8                                   \\
Linear Probe                         & 80.0                     & \multicolumn{1}{c|}{67.8}               & 80.5                     & \multicolumn{1}{c|}{66.8}               & 82.1                     & \multicolumn{1}{c|}{68.2}               & 83.5                     & 73.6                                   \\
Continual-CLIP  & 74.0                     & \multicolumn{1}{c|}{66.7}               & 75.1                     & \multicolumn{1}{c|}{66.7}               & 75.9                     & \multicolumn{1}{c|}{66.7}               & 76.5                     & 66.7                                   \\ \midrule
Ours                                 & \bf{83.9$\pm$0.24}       & \multicolumn{1}{l|}{\bf{77.6$\pm$0.13}} & \bf{84.2$\pm$0.23}       & \multicolumn{1}{l|}{\bf{76.9$\pm$0.38}} & \bf{84.4$\pm$0.33}       & \multicolumn{1}{l|}{\bf{76.3$\pm$0.69}} & \bf{84.8$\pm$0.22}       & \multicolumn{1}{l}{\bf{76.1$\pm$0.32}} \\ \bottomrule
\end{tabular}%
}
\caption{Comparison of our approach with the state-of-the-art on CIFAR100 B0 in detail.}
\label{tab:my-table-b0}
\end{table*}

\begin{table*}[]
\centering
\begin{tabular}{r|cccc}
\toprule
\multirow{4}{*}{Method} & \multicolumn{4}{c}{CIFAR-100 B50}                                                                                                                                           \\ \cline{2-5} 
                        & \multicolumn{4}{c}{\# tasks}                                                                                                                                                \\
                        & \multicolumn{2}{c|}{5}                                                                         & \multicolumn{2}{c}{10}                                                     \\ \cline{2-5} 
                        & \multicolumn{1}{c|}{Avg}            & \multicolumn{1}{c|}{Last}                                & \multicolumn{1}{c|}{Avg}             & Last                                \\ \midrule
iCaRL                   & 65.1                                & \multicolumn{1}{c|}{56.0}                                & 58.6                                 & 49.5                                \\
LUCIR                   & 64.3                                & \multicolumn{1}{c|}{52.7}                                & 59.9                                 & 48.2                                \\
BiC                     & 66.6                                & \multicolumn{1}{c|}{55.1}                                & 60.3                                 & 48.7                                \\
WA                      & 64.0                                & \multicolumn{1}{c|}{52.8}                                & 57.9                                 & 48.1                                \\
PODNet                  & 67.3                                & \multicolumn{1}{c|}{55.9}                                & 64.0                                 & 51.7                                \\
DER                     & 73.2                                & \multicolumn{1}{c|}{66.0}                                & 72.8                                 & 65.6                                \\
DyTox                   & -                                   & \multicolumn{1}{c|}{-}                                   & -                                    & -                                   \\ \midrule
L2P                     & 77.9                                & \multicolumn{1}{c|}{68.8}                                & 76.8                                 &  69.7                                \\
Linear Probe            & 79.8                                & \multicolumn{1}{c|}{66.8}                                & 80.7                                 & 68.2                                \\
Continual-CLIP          & 69.7                                & \multicolumn{1}{c|}{66.7}                                & 69.5                                 & 66.7                                \\ \midrule
Ours                    & \bf{81.4$\pm$0.33} & \multicolumn{1}{c|}{\bf{77.0$\pm$0.86}} & \bf{82.9$\pm$1.51} & \bf{76.4$\pm$1.00} \\ \bottomrule
\end{tabular}
\caption{Comparison of our approach with the state-of-the-art on CIFAR100 B50 in detail.}
\label{my_table_b50}
\end{table*}

\vspace{7em}

\newpage
\appendix

\section{Implementation Details}

All Experiments were conducted on an Ubuntu workstation equipped with dual 3090 GPUs, 10 CPUs, and 16GB of peak memory usage. PyTorch version 1.13 with CUDA version 11.8 eas used.

\begin{table*}
\centering
{%
\begin{tabular}{r|cccccc}
\toprule
\multirow{4}{*}{Method} & \multicolumn{6}{c}{ImageNet-100 B0}                                                                                                                      \\ \cline{2-7} 
                        & \multicolumn{6}{c}{\# tasks}                                                                                                                             \\
                        & \multicolumn{2}{c|}{5}                               & \multicolumn{2}{c|}{10}                              & \multicolumn{2}{c}{20}                     \\ \cline{2-7} 
                        & \multicolumn{1}{c|}{Avg} & \multicolumn{1}{c|}{Last} & \multicolumn{1}{c|}{Avg} & \multicolumn{1}{c|}{Last} & \multicolumn{1}{c|}{Avg}       & Last      \\ \midrule
iCaRL                   & 78.1                     & \multicolumn{1}{c|}{65.2} & 74.1                     & \multicolumn{1}{c|}{58.5} & \multicolumn{1}{c|}{69.0}      & 50.9      \\
End2End                 & 75.5                     & \multicolumn{1}{c|}{64.0} & 70.1                     & \multicolumn{1}{c|}{50.3} & \multicolumn{1}{c|}{68.3}      & 48.9      \\
LUCIR                   & 76.0                     & \multicolumn{1}{c|}{64.0} & 70.5                     & \multicolumn{1}{c|}{55.3} & \multicolumn{1}{c|}{64.7}      & 47.8      \\
PODNet                  & 78.2                     & \multicolumn{1}{c|}{66.2} & 72.3                     & \multicolumn{1}{c|}{57.2} & \multicolumn{1}{c|}{66.7}      & 48.9      \\
LUCIR-DDE               & 77.2                     & \multicolumn{1}{c|}{65.8} & 71.7                     & \multicolumn{1}{c|}{56.8} & \multicolumn{1}{c|}{66.2}      & 40.0      \\
RM                      & 75.5                     & \multicolumn{1}{c|}{62.2} & 70.4                     & \multicolumn{1}{c|}{53.2} & \multicolumn{1}{c|}{65.4}      & 45.7      \\
DER                     & -                        & \multicolumn{1}{c|}{-}    & 77.2                     & \multicolumn{1}{c|}{66.7} & \multicolumn{1}{c|}{-}         & -         \\ \midrule
L2P                     & 81.3                     & \multicolumn{1}{c|}{\bf{76.3}} & 80.1                     & \multicolumn{1}{c|}{75.8} & \multicolumn{1}{c|}{80.7}      & 76.1      \\
Linear Probe            & 79.4                     & \multicolumn{1}{c|}{64.5} & 81.8                     & \multicolumn{1}{c|}{67.4} & \multicolumn{1}{c|}{84.0}      & 72.0      \\
Continual-CLIP          & 81.7                     & \multicolumn{1}{c|}{75.3} & 83.1                     & \multicolumn{1}{c|}{75.3} & \multicolumn{1}{c|}{83.5}     & 75.3      \\ \midrule
Ours                    & \bf{84.9$\pm$0.68}                & \multicolumn{1}{c|}{74.7$\pm$0.76} & \bf{85.3$\pm$0.92}                & \multicolumn{1}{c|}{\bf{76.8$\pm$0.21}} & \multicolumn{1}{c|}{\bf{84.1$\pm$1.11}} & \bf{77.0$\pm$0.15} \\ \bottomrule
\end{tabular}%
}
\caption{Comparison of our approach with the state-of-the-art on ImageNet100 B0 in detail.}
\label{tab:my-table-img-b0}
\end{table*}

\begin{table*}[]
\centering
{%
\begin{tabular}{r|cccc|c}
\toprule
\multirow{4}{*}{Method} & \multicolumn{4}{c|}{ImageNet-100 B50}                                                                       & ImageNet-R B0 \\ \cline{2-6} 
                        & \multicolumn{4}{c|}{\# tasks}                                                                               & tasks         \\
                        & \multicolumn{2}{c|}{5}                               & \multicolumn{2}{c|}{10}                              & 20            \\ \cline{2-6} 
                        & \multicolumn{1}{l|}{Avg} & \multicolumn{1}{l|}{Last} & \multicolumn{1}{l|}{Avg} & \multicolumn{1}{l|}{Last} & Last          \\ \midrule
iCaRL                   & 60.7                     & \multicolumn{1}{c|}{44.7} & 57.3                     & 44.4                      & -             \\
End2End                 & 61.0                     & \multicolumn{1}{c|}{52.1} & 58.5                     & 52.2                      & -             \\
LUCIR                   & 77.2                     & \multicolumn{1}{c|}{68.2} & 66.9                     & 56.8                      & -             \\
PODNet                  & 80.3                     & \multicolumn{1}{c|}{73.5} & 79.0                     & 70.8                      & -             \\
LUCIR-DDE               & 78.8                     & \multicolumn{1}{c|}{68.1} & 68.4                     & 57.9                      & -             \\
RM                      & 56.9                     & \multicolumn{1}{c|}{41.8} & 57.7                     & 37.3                      & -             \\
DER                     & -                        & \multicolumn{1}{c|}{-}    & 78.2                     & 74.9                      & 66.7          \\ \midrule
L2P                     & 74.8                     & \multicolumn{1}{c|}{74.1} & 71.9                     & 72.5                      & 74.1          \\
Linear Probe            & 81.4                     & \multicolumn{1}{c|}{67.2} & 83.1                     & 72.0                      & 57.5          \\
Continual-CLIP          & 79.2                     & \multicolumn{1}{c|}{75.3} & 79.3                     & 75.3                      & 72.0          \\ \midrule
Ours                    & \bf{85.1$\pm$1.50}                & \multicolumn{1}{c|}{\bf{76.8$\pm$0.34}} & \bf{85.4$\pm$1.59}                & \bf{77.4$\pm$0.26}                 & \bf{75.9$\pm$0.82}     \\ \bottomrule
\end{tabular}%
}
\caption{Comparison of our approach with the state-of-the-art on ImageNet100 B50 and ImageNet-R B0 in detail.}
\label{tab:my-table-img-b50}
\end{table*}

\section{More Experiment Results}
In this section, we provide comprehensive experimental results on CIFAR100, ImageNet100, ImageNet-R, and ImageNet-Full across various settings and distinct random seeds (class orders). The number after "$\pm$" signifies the standard deviation over multiple experiments. 

\subsection{Evaluation on CIFAR100}
As shown in Table~\ref{tab:my-table-b0} and Table~\ref{my_table_b50}, our approach achieves significantly better performance compared to the state-of-the-art. When compared to conventional methods of learning from scratch, our approach improves over the best competing methods DER and DyTox by a large margin on both average and last task metrics in both the B0 and B50 settings. Our model is able to consistently maintain relatively high performance across subsequent tasks. Of course, such high performance is partially attributable to our use of the pre-trained CLIP model. However, compared to the simple zero-shot application of CLIP to continual learning, such as Linear Probe or Continual-CLIP, our approach still exhibits significant advantages.

Notably, the Linear Probe results are higher than those of Continual-CLIP due to the use of incoming data and the exemplars stored from the previous tasks.

Furthermore, even with the incorporation of a 2000-exemplar set, the improvement of L2P over Continual-CLIP (zero-shot) is insignificant. This indicates that the effectiveness of the prompt pool used in the original L2P is largely due to the supervised pre-training of the model.

In Table~\ref{tab:adaptor} we see that our method (just adding one linear adapter layer) can outperform others. Specifically, in the last task, our method (Ours w/o R) is at least 1.8 points higher than all attention methods. Furthermore, using the parameter retention strategy (Ours) further improves accuracy.

\begin{table*}
\centering

{%
\begin{tblr}{
  vline{2} = {-}{},
  hline{1-2,4} = {-}{},
  cells = {c}
}
Adapter & Additional Block &  Self-attention Adapter   &Finetuning Last & Ours (w/o R)&  Ours   \\
Avg      & 81.1     & 75.8     &82.9 &83.2& 84.2\\
Last     & 68.9     & 71.4    &71.4 & 73.2 & 76.9 
\end{tblr}
}
\caption{Performance of our method with different types of attention in terms of average (Avg) and last-task accuracy (Last) in the CIFAR-100 B0-10 setting.}
\label{tab:adaptor}
\end{table*}

\begin{table*}
\centering

{%
\begin{tblr}{
  vline{2} = {-}{},
  hline{1-2,4} = {-}{},
  cells = {c}
}
Methods & iCaRL & WA & DER  &DER(w/o P)&   Ours   \\
Avg      & 38.4     & 65.7    & 66.7 &68.8& 75.6\\
Last     & 22.7     & 55.6    & 58.6 &60.2&  63.8
\end{tblr}
}
\caption{Comparison of our approach with the state-of-the-art on ImageNet-Full.}
\label{tab:fullimagenet}

\end{table*}

\subsection{Evaluation on ImageNet-100 and Imagenet-R}
In Tables~\ref{tab:my-table-img-b0}~and~\ref{tab:my-table-img-b50} we see that, in the B50 setting, our outcomes consistently surpass the baselines. In the B0 setting, our average performance also slightly exceeds them. It is important to note that the average results for Continual-CLIP are derived from three experiments with the same seed and shuffled class orders, which could potentially lead to a marginally lower value compared to those reported in the original paper.

\subsection{Evaluation on ImageNet-Full}
In this section, we follow DER and  evaluate on ImageNet-Full with 1000 classes. As shown in Table~\ref{tab:fullimagenet}, in B0-10 setting (10 tasks with 100 classes each), our method improves DER by about 3.6\% at the last task and about 6.8\% on average over all tasks.

\subsection{Attention Block}
Here, we discuss several different attention mechanisms (see Table~\ref{tab:adaptor}). First, we can adjust the last self-attention block in the ViT transformer of CLIP  during training without any adapter (Finetuning Last). Second, we can add another self-attention block after the transformer and only update this module without updating the pre-trained model (Additional Block). Finally, as mentioned in our main text (Self-attention Adapter), we add self-attention to the 512-dimensional features output by the ViT transformer.

\end{document}